\documentclass[letterpaper, 10 pt, conference]{ieeeconf}  % Comment this line out if you need a4paper

\IEEEoverridecommandlockouts                              % This command is only needed if 
\usepackage{times}
\usepackage{epsfig}
\usepackage{graphicx}
\usepackage{amsmath}
\usepackage{amssymb}
\let\labelindent\relax
\usepackage{enumitem}
\usepackage{multirow}
\usepackage{makecell}
\usepackage{float}
\usepackage{subcaption}
\usepackage[dvipsnames]{xcolor}
\usepackage[breaklinks=true,bookmarks=false]{hyperref}
\usepackage[space]{cite} %space option makes the citations to be nicely combined

\newcommand{\norm}[1]{\left\lVert#1\right\rVert}

\newcommand{\sevensensesymbol}{1}
\newcommand{\aslsymbol}{2}

\title{\LARGE \bf
Fast Image-Anomaly Mitigation for Autonomous Mobile Robots
} 

\begin{document}

\author{Gianmario Fumagalli$^{\sevensensesymbol,}$$^{\aslsymbol}$, Yannick Huber$^{\sevensensesymbol}$, Marcin Dymczyk$^{\sevensensesymbol}$, Roland Siegwart$^{\aslsymbol}$, Renaud Dub\'e$^{\sevensensesymbol}$\\\small$^{\sevensensesymbol}$Sevensense Robotics AG, {\tt\footnotesize \{firstname.lastname\}@sevensense.ch}
\\
$^{\aslsymbol}$Autonomous Systems Lab, ETH Z\"{u}rich, {\tt\footnotesize \{firstname.lastname\}@mavt.ethz.ch}}

\maketitle
\thispagestyle{empty}
\pagestyle{empty}

%%%%%%%%%%%%%%%%%%%%%%%%%%%%%%%%%%%%%%%%%%%%%%%%%%%%%%%%%%%%%%%%%%%%%%%%%%%%%%%%
\begin{abstract}
Camera anomalies like rain or dust can severely degrade image quality and its related tasks, such as localization and segmentation.   
In this work we address this important issue by implementing a pre-processing step that can effectively mitigate such artifacts in a real-time fashion, thus supporting the deployment of autonomous systems with limited compute capabilities.
We propose a shallow generator with aggregation, trained in an adversarial setting to solve the ill-posed problem of reconstructing the occluded regions.
We add an enhancer to further preserve high-frequency details and image colorization.
We also produce one of the largest publicly available datasets\footnote{\url{https://github.com/sevensense-robotics/image_anomalies_dataset}} to train our architecture and use realistic synthetic raindrops to obtain an improved initialization of the model.
We benchmark our framework on existing datasets and on our own images obtaining state-of-the-art results while enabling real-time performance, with up to 40x faster inference time than existing approaches.
\end{abstract}

%%%%%%%%%%%%%%%%%%%%%%%%%%%%%%%%%%%%%%%%%%%%%%%%%%%%%%%%%%%%%%%%%%%%%%%%%%%%%%%%
\vspace{0.25cm}
\section{INTRODUCTION} % 600 words
\begin{figure}[t]
\begin{center}
    \includegraphics[scale=0.1575]{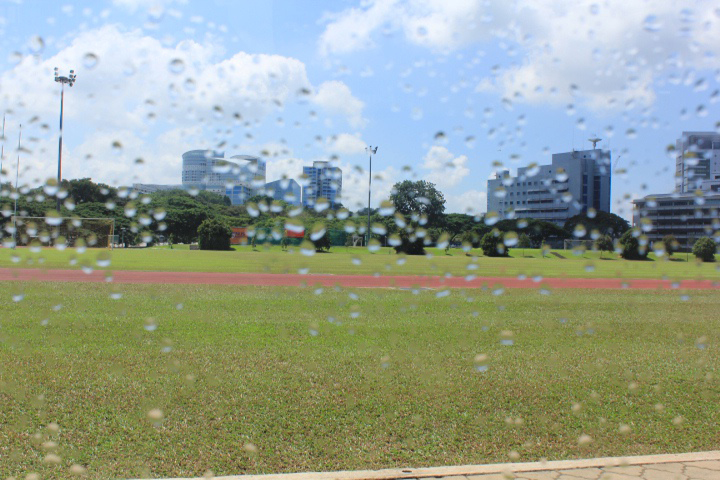}
    \includegraphics[scale=0.1575]{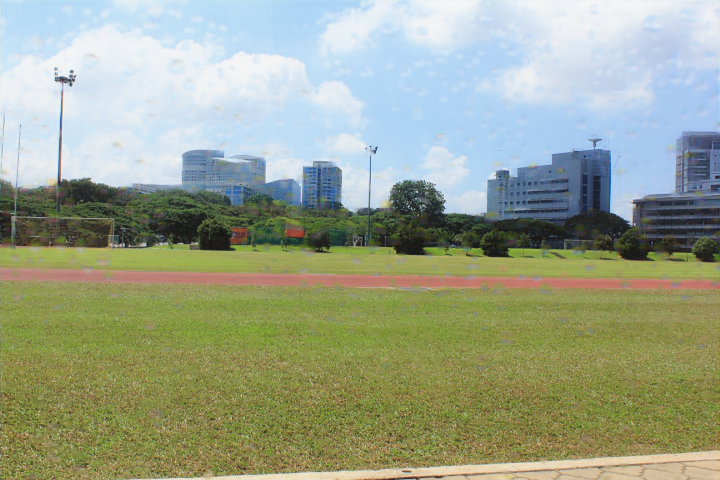}
    \includegraphics[scale=0.315]{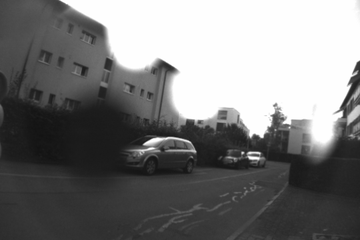}
    \includegraphics[scale=0.315]{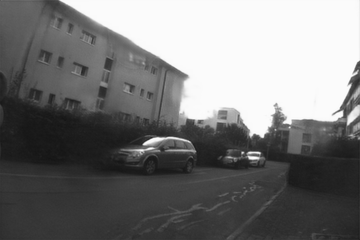}
    \includegraphics[scale=0.315]{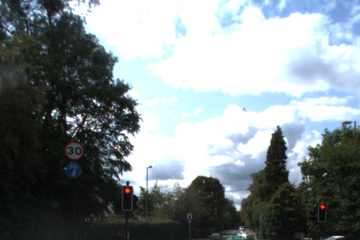}
    \includegraphics[scale=0.315]{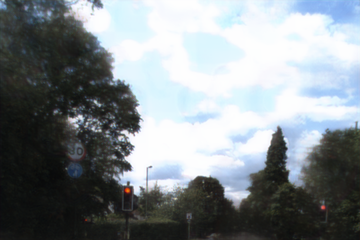}
\end{center}
\caption{Fast camera-anomaly mitigation produced by our GAN architecture. Rain-affected samples (left column) are fed into our generator that, alongside the enhancer module, learns to restore the original clean image (right column) in a real-time fashion.}
\vspace{-0.5cm}
\label{img:intro}
\end{figure}

With the continuous development of autonomous robots, their deployment outside labs is rapidly increasing.%~\cite{harel2020autonomics}.
While representing a great step towards the future, leaving machines operate in a less-controlled, less-predictable environment poses completely new challenges.
Particularly crucial are the ones related to the so-called camera anomalies, which are impurities that can affect the camera lenses and deteriorate the captured images, as shown in Figure~\ref{img:intro}, left column.
For instance, the vision system of robots moving in outdoor terrains can be spoiled by lifted soil, dust and by the weathering, rain especially~\cite{gu2009removing}.
Raindrops on lenses have been largely the most studied camera anomaly~\cite{roser2009video, roser2010realistic, you2015adherent}, due to the higher probability to occur and the wider impact they have on the image.
Indeed, their rounded shapes lead to a fish-eye-lens effect, that refracts the light in a different way with respect to the background and displays different -- sometimes far -- parts of the scene. 
Considering that the focal point is usually at infinity, with the unaffected regions being in focus and the raindrops looking extremely blurred, it is not surprising that the degradation of the image would be consistent, thus reducing dramatically the performance of vision-related tasks, such as classification and segmentation.
%
%@Gianmario when you reshuffle this part please remember that your main contribution is the novel light-weight real-time applicable architecture. The data part is nice but I would only see the dataset collection as a secondary contribution (aka strong experiment component) and wouldn't compromise the focus on the real-time aspect (it's also not clear if we want to make it public)

Although some research has moved towards building models that can adapt to uncertain conditions and challenging environments~\cite{valada2019self}, DeRaindrop~\cite{qian2018attentive} and Porav \textit{et al.}~\cite{porav2019can} have experimentally showed that a de-noising pre-processing step is more effective.
These architectures are able to produce realistic de-rained samples, but at the cost of not being real-time capable on platforms with limited computing power, like autonomous robots. 
Following their steps, our approach towards the task aims at developing an effective and efficient pre-processing step able to restore the original clean image, while enabling real-time performance.
We developed a light-weight generator adapted from Porav \textit{et al.}~\cite{porav2019can} that estimates the occluded regions, coupled with an enhancer module~\cite{qu2019enhanced} that preserves input colors and details.
Our architecture is trained in an adversarial setting~\cite{goodfellow2014generative} with a discriminator~\cite{li2016precomputed} that has proven effective for this task~\cite{qian2018attentive, porav2019can}.
%These works provide for raindrops adherent to the camera lens. 
%their performance is particularly limited by the cost of producing high-quality \textit{rainy} samples for every task, which sum up to the already expensive \textit{clean} data collection.
%
With our framework, we can compete with state-of-the-art methods on datasets from~\cite{qian2018attentive, porav2019can} and outperform all existing methods with similar results by a large margin in terms of inference speed.
In addition, we trained the model on our own dataset captured with the novel sensor shown in Figure~\ref{img:Sensor}.
We collected the largest available dataset for the task, comprising of pairs of anomaly-affected and clean images, as described in Section~\ref{sec_dataset}. 
The majority of samples is degraded by raindrops adherent on glass and a few are spoiled with dust and dirt to show how our architecture can be adapted to clean images affected by other, less researched anomalies.
Alongside with it, we provide a realistic synthetic drop generation that mimics the physical photometric effect of raindrops on camera lens.
With respect to our reference model~\cite{porav2019can}, we improved by creating proto-raindrops that are also full-fledged out of focus.
A foreground layer of computer-generated raindrops can be applied to huge existing datasets, like ImageNet~\cite{deng2009imagenet} or MS-COCO~\cite{lin2014microsoft}, to obtain cost-free rainy samples.
We used our synthetic images in support of real data, as a pre-training step to produce a precise initialization for the weights before the actual training.
Section~\ref{syndata} exposes the process of creating our synthetic raindrops and displays their use for our purposes.
To summarize, our main contributions are:
\begin{itemize}
    \item A light architecture that can perform anomaly mitigation with state-of-the-art performance in real-time;
    \item A realistic synthetic-drop-generation technique that outputs rainy images with out-of-focus raindrops\footnote{A video demonstration is available at \url{https://youtu.be/R50peOmqCj4}.};
    \item The largest dataset for the task of removing camera-attached anomalies, with rainy or dusty images with the corresponding clean ground truth.
\end{itemize}
\section{RELATED WORK}
Several attempts have been done throughout the years to reduce the effect of camera anomalies, such as snow, fog, haze, dirt, rain streaks and raindrops, with both deep-learning (DL) and non-deep-learning techniques.
\subsection{Non-DL methods}
Some researchers in this field, especially in early works, have tackled the problem using a non-deep-learning approach. 
Their main limitation is that they all require high-framerate videos instead of single images, thus reducing their applicability in a real-world scenario.
%
%In~\cite{zhang2006rain} the authors implement a rain removal algorithm that leverages both temporal and chromatic properties of pixels across a video sequence, stating that a pixel is never covered by rain throughout the entire video and the changes in RGB values are approximately the same. It also has the limitation that it only works for static cameras.
%
Roser and Geiger~\cite{roser2009video} developed a rain-detection and image-reconstruction algorithm using photometric properties (optical path) applied
to a spherical raindrop model, which is fitted multiple times in the image to detect raindrop regions. Image reconstruction is done evaluating the intensity level of pixels near raindrops. In their subsequent work~\cite{roser2010realistic} they change the raindrop model to cubic Bezier Curves, which is more realistic. However, the range of shapes and sizes of raindrops is too high to be adequately fitted.
In~\cite{kim2014stereo} stereo video sequences are needed to perform spatio-temporal frame warping with a median filter across three selected frames to reconstruct the clean image, while in~\cite{you2015adherent} raindrops are detected and removed by looking at local properties, i.e. spatial derivatives and changes in optical flow.
Others~\cite{kim2015video, ren2017video} model background fluctuations with a multi-label Markov field and reconstruct the image via a low-rank representation (SVD decomposition) of the background.
The latest work by~\cite{jiang2018fastderain} applies a split augmented Lagrangian Shrinkage algorithm to directional and temporal gradients to efficiently remove rain streaks in videos.
\subsection{DL methods}
Latest research is mainly focusing on deep-learning techniques for their capacity to store information during training~\cite{burger2012image} allowing them to perform anomaly mitigation on single images.
Apart from the work by Eigen \textit{et al.}~\cite{eigen2013restoring}, where a Convolutional Neural Network (CNN) is used to restore an image taken through a window covered by rain or dirt, the great majority of deep-learning techniques involve the use of Generative Adversarial Networks~\cite{goodfellow2014generative}, as they have proved successful in several image-to-image translation problems~\cite{isola2017image, zhu2017unpaired}.
%
%Image de-noising is a very studied topic, but main attention is given to image dehazing
Severeal works in this field address the task of image de-noising applied to de-hazing~\cite{cai2016dehazenet, ren2016single, li2017aod, zhang2018densely, ren2018gated, yang2018proximal, liu2019griddehazenet} and rain streak deletion~\cite{zhang2019image, ren2019progressive, wei2019coarse, wang2019spatial, chen2019gated} and can be related to the equally important topic of camera-anomaly mitigation.
Qian \textit{et al.}~\cite{qian2018attentive} inject visual attention in both generator and discriminator architectures to focus on raindrop regions. The probabilities of belonging to a raindrop are estimated using a recurrent neural network~\cite{rumelhart1985learning, jordan1997serial} alongside with an LSTM module~\cite{hochreiter1997long}. The remaining part of the generator consists of a convolutional autoencoder, and the discriminator is represented by a CNN. The authors train the architecture on their static dataset and produce state-of-the-art results at the cost of slow inference time, computational complexity and complex training scheme due to the attention module.
Porav \textit{et al.}~\cite{porav2019can} remove the attention and develop a simpler architecture similar to~\cite{wang2018high}. The generator employs the autoencoder structure with skip-connection layers to preserve input structure and details. It is trained in an adversarial setting alongside a Patch-GAN discriminator~\cite{li2016precomputed}. They also published a dataset that allowed us to compare our results with theirs. Again, while being simpler, their architecture is still heavy and does not ensure real-time performance, which is critical for real-world robotic applications.
\begin{figure*}[t]
    \includegraphics[width = \textwidth]{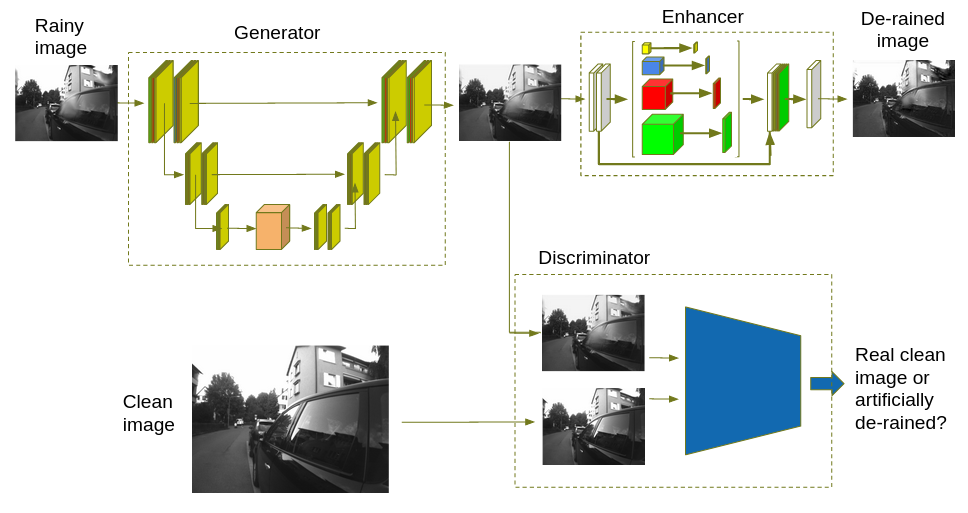}
\caption{Our full pipeline for anomaly removal. The affected image is first run through the generator that performs the first mitigation. Then, after concatenation with the original image, it is fed into the enhancer to produce the final output.}
\vspace{-0.5cm}
\label{img:Architecture}
\end{figure*}
\section{METHODOLOGY}
%Images:
%\begin{itemize}
%    \item Image of whole architecture;
%    \item Maybe zoom of enhancer module;
%    \item Samples of synthetic raindrops.
%\end{itemize}
%\subsection{Physical phenomenon or baseline?}
%\begin{itemize}
%    \item Short section where we present the raindrop image formation problem, similar to sec 3 of %\cite{qian2018attentive};
%    \item Talk about the raindrops as a binary mask applied to the image that has to be detected and removed;
%    \item Maybe also briefly about light scattering that can affect both raindrop regions and background %brightness.
%\end{itemize}
\subsection{Architecture}\label{sec_arc}
We address the task of image de-raining as an image-to-image translation problem, where rainy and clear images are regarded as two different image styles. Our architecture comprises a generator and an enhancer. The former tries to fool the discriminator during training by producing fake de-rained images in an adversarial fashion; the latter aims to improve the resulting images by preserving input color and details. Figure~\ref{img:Architecture} displays our full architecture.
\begin{figure}[h]
\begin{center}
    \includegraphics[scale=0.33]{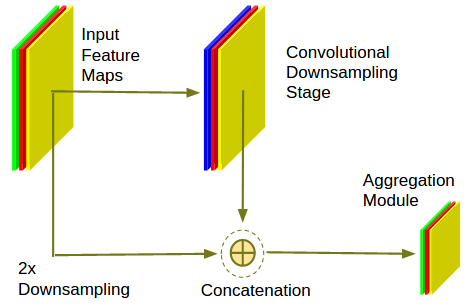}
    \end{center}
    \caption{Zoom of the aggregation module.}
    \vspace{-0.5cm}
    \label{img:aggregation}
\end{figure}
{\noindent\textbf{Generator}\quad}\label{sec_gen}
%\begin{itemize}
%    \item Describe the generator architecture by looking at papers \cite{isola2017image, %wang2018high, porav2019can} because their generator is similar.
%\end{itemize}
Our generator architecture is inspired from Pix2Pix HD~\cite{wang2018high} and from~\cite{porav2019can}, although much lighter. It has an encoder-decoder structure with a sequence of residual blocks~\cite{he2016deep} in between. In contrast to the others, our encoder employs only two downsampling convolutional layers, with 16 and 32 filters respectively, thus drastically reducing the number of parameters. To refine the resulting feature maps we added a convolutional aggregation module after each downsampling stage, with the corresponding number of filters. The aggregation is performed between the input and the output of the downsampling convolutional layer, where the input is downsampled to match the output, as showed in Figure~\ref{img:aggregation}.
The encoder outputs feature maps with a quarter of the original size, that are then fed into a series of 9 residual blocks, with 64 filters each, that perform the actual image restoration, focusing on raindrops and trying to estimate the occluded regions. Its output is passed through a decoder with a similar structure as the encoder, with the only difference that skip-connection layers are added as input to the aggregation blocks to preserve useful input information.
{\noindent\textbf{Discriminator}\quad}\label{sec_disc}
%\begin{itemize}
%    \item Describe the enhancer by looking at \cite{li2016precomputed}.
%\end{itemize}
Our discriminator helps the generator to output more realistic results during training~\cite{goodfellow2014generative}. It is similar to PatchGAN~\cite{li2016precomputed}, which has proven successful for this task~\cite{isola2017image, wang2018high, porav2019can}. We use 3 convolutional layers, enforcing the discriminator to classify 14x14 patches as real or fake, instead of the whole image, thus helping the generator to produce more detailed outputs.
{\noindent\textbf{Enhancer}\quad}\label{sec_enh}
%\begin{itemize}
%    \item Describe the enhancer by looking at \cite{qu2019enhanced}.
%\end{itemize}
%
Reducing the size of the network comes at the cost of reduced performance, especially in terms of details and colorization. To overcome this issue, we leverage the enhancer module recently proposed by Qu \textit{et al.}~\cite{qu2019enhanced}. Although originally introduced for image de-hazing, it can still be applied to image de-raining, as proved in our experiments. It contains a 4-scale pyramid pooling that processes feature maps at resolutions reduced by factors of 4, 8, 16 and 32 in parallel, which allows the model to capture information at different levels of detail. A 1x1 convolution is applied to each of them to weight the channels adaptively. After up-sampling to the original resolution, they are concatenated with the original feature maps using a 3x3 convolution to produce the final outputs.

\begin{figure*}[t]
\begin{center}
    \includegraphics[scale=0.22]{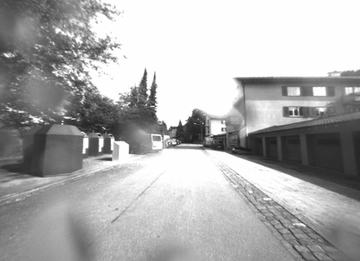} % first figure itself
    \includegraphics[scale=0.22]{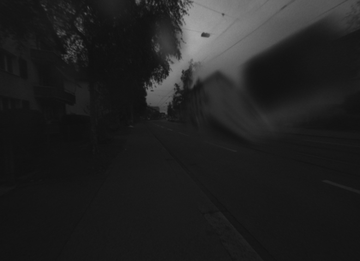} % first figure itself
    \includegraphics[scale=0.22]{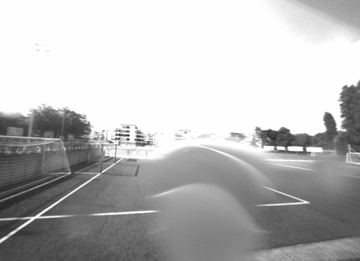} % first figure itself
    \includegraphics[scale=0.22]{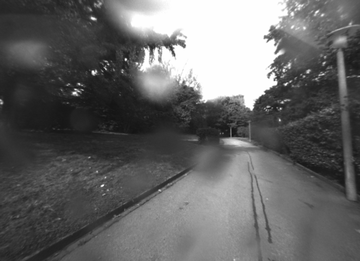} % first figure itself
    \includegraphics[scale=0.22]{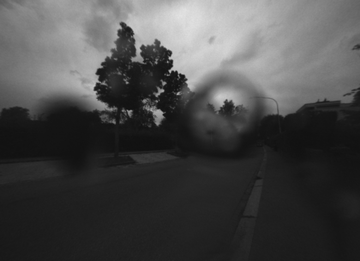} % first figure itself
    \includegraphics[scale=0.22]{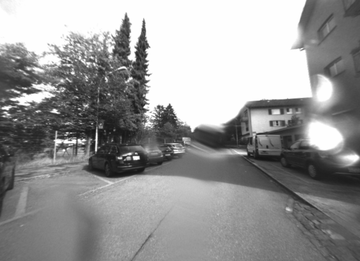} % first figure itself
\end{center}
    \caption{Samples from our dataset.}
    \vspace{-0.3cm}
    \label{img:dataset}
\end{figure*}
{\noindent\textbf{Losses}\quad}\label{sec_loss}
%\begin{itemize}
%    \item Talk about the losses we use;
%    \item GAN loss;
%    \item Multi-scale discriminator loss;
%    \item Perceptual loss;
%    \item Pixel loss.
%\end{itemize}
Our objective function combines several losses to take into account the different qualities we require for our outputs. The first two are applied to the generator and discriminator and preserve global context information, while the last two work on the enhancer and help with details and colors. The \textit{adversarial loss}~\cite{goodfellow2014generative} is defined as
\vspace{-0.05cm}
\begin{equation}\label{eqGAN}
    \mathcal{L}_{GAN} = (1 - D(G(\mathcal{A})))^{2},
    \vspace{-0.05cm}
\end{equation}
where $\mathcal{A}$ stands for an image affected by anomalies, $D$ for discriminator and $G$ for generator. It is employed in the interplay between generator and discriminator, where the former tries to fool the latter by producing realistic results. For this reason, it has to be applied directly to the output of the generator, and not on the enhanced results. The same happens with the \textit{feature matching loss}~\cite{wang2018high}
% \vspace{-0.15cm}
\begin{equation}\label{eqFM}
    \mathcal{L}_{FM} = \sum_{n=1}^{n_{FM}}\frac{\norm{D(\mathcal{C})_{n}-D(G(\mathcal{A}))_{n}}_1}{2^{n_{FM}-n}}.
    % \vspace{-0.15cm}
\end{equation}
Here, $n_{FM}$ stands for the number of selected discriminator layers and $\mathcal{C}$ for the clean image. This loss penalizes the distance between the intermediate features extracted by the discriminator from both the ground-truth and the reconstructed images. This way, it ensures that the generator outputs fake samples that are close to reality even at multiple scales. On the other hand, the \textit{perceptual loss}~\cite{johnson2016perceptual}
\vspace{-0.05cm}
\begin{equation}\label{eqVGG}
    \mathcal{L}_{VGG} = \sum_{n=1}^{n_{VGG}}\frac{\norm{VGG(\mathcal{C})_{n}-VGG(E(G(\mathcal{A})))_{n}}_1}{2^{n_{VGG}-n}}
    \vspace{-0.05cm}
\end{equation}
is applied between enhanced and clean image. It computes the absolute error between activations of neurons at certain layers of a VGG network~\cite{simonyan2014very} pretrained on ImageNet~\cite{deng2009imagenet}. By doing so, we encourage the network to preserve high-level perceptual features when restoring the image. The \textit{fidelity loss}
\vspace{-0.15cm}
\begin{equation}\label{eqFID}
    \mathcal{L}_{FID} = \norm{\mathcal{C}-E(G(\mathcal{A}))}_2
    \vspace{-0.05cm}
\end{equation}
measures the L2 pixel-wise difference between the enhanced output and the clean image.
The simplicity of our network allows the minimization of the sum of all these losses in a unique step, which differs from~\cite{qu2019enhanced} where a complex training scheme is proposed. Thus, the objective function to be minimized during training is:
\vspace{-0.05cm}
\begin{equation}\label{eqG}
    \mathcal{L}_{G} = \mathcal{L}_{GAN}+\mathcal{L}_{FM}+\mathcal{L}_{VGG}+\mathcal{L}_{FID}.
    % \vspace{-0.15cm}
\end{equation}

\subsection{Synthetic Data}\label{syndata}
\begin{figure}[t]
\begin{center}
    \includegraphics[scale=0.68]{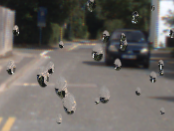}
    \includegraphics[scale=0.68]{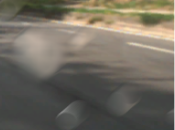}
\end{center}
\caption{Differences between our synthetic raindrops (right) and our reference model~\cite{porav2019can} (left). Ours are out of focus and with different brightness with respect to the background, which is more common for fixed-focus cameras used for mobile robotics.}
\vspace{-0.3cm}
\label{img:Syn}
\end{figure}

%\begin{itemize}
%    \item Short section, mainly our syn-raindrops are similar to the ones in \cite{porav2019can};
%    \item Difference 1: our raindrops are out of focus;
%    \item Difference 2: we consider change of image brightness inside raindrop (more realistic);
%    \item If we want to add dust/dirt explain the process of dirt computer-generation;
%    \item Here or in experiments (maybe add a table of with and without synthetic raindrops initialization): the use we do of synthetic raindrops.
%\end{itemize}
We model our synthetic drops similar to Porav \textit{et al.}~\cite{porav2019can}. However, we noticed that their syn-drops are in focus and with sharp borders, in contrast to what happens in reality. To synthesize out-of-focus raindrops we introduce multiple shiftings in random directions where our camera anomalies are situated. Moreover, we increase the brightness inside some random raindrop regions, as we experienced when recording our real dataset. An example of our improvements with respect to our reference are displayed in Figure~\ref{img:Syn}.
Images corrupted with our computer-generated raindrops are used to train the network and the resulting weights are re-used as initialization for the training on our real dataset, to provide a task-related initial configuration. If applied with this modality, our synthetic raindrops proved to be more realistic than the ones produced by our reference model~\cite{porav2019can}. Please refer to Section~\ref{exp_syn} for a quantitative evaluation.

\vspace{0.2cm}
\section{EXPERIMENTS}
\subsection{Dataset}\label{sec_dataset}
%\begin{itemize}
%    \item Image of sensor;
%    \item Samples from our dataset (multiple small images);
%    \item Explain data-recording process;
%    \item Image undistortion, registration and cropping;
%    \item Size of images;
%    \item Number of data and train, val and test splits.
%\end{itemize}

\begin{figure}[t]
\begin{center}
    \includegraphics[scale=0.23]{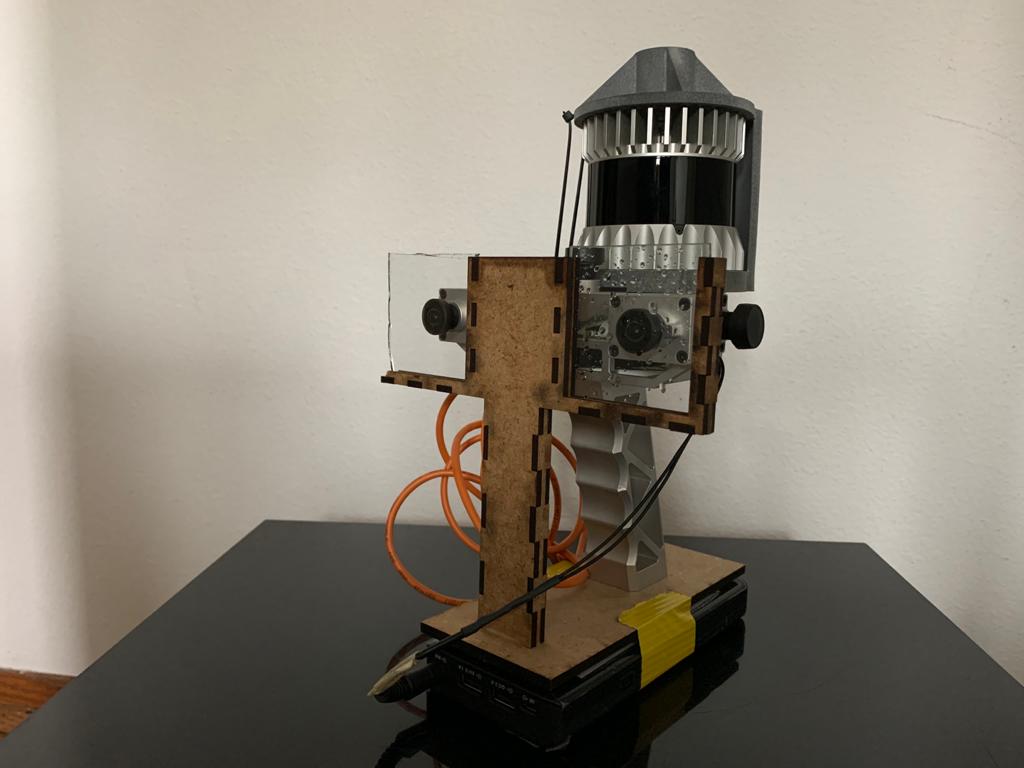}
\end{center}
\caption{Setup of our device for data collection. A wooden frame allows to place a glass panel in front of each of the two anterior cameras of the Alphasense Core multi-camera sensor. One is kept clean and the other one is corrupted with anomalies.}
\vspace{-0.5cm}
\label{img:Sensor}
\end{figure}

Here we present the device used to collect our dataset, shown in Figure~\ref{img:Sensor}. We used the Alphasense Core multi-camera sensor\footnote{\url{https://github.com/sevensense-robotics/alphasense_core_manual}} manufactured by Sevensense Robotics. The two front cameras (Sony IMX-287) are spaced by a baseline of 140 mm and are synchronized to record images of the same scene, to ensure we have both the ground-truth clean image and the affected one. Instead of directly spoiling one camera lens, we have built a frame to sustain two pieces of glass, one placed in front of each camera. This way, we could affect only one of the two cameras with the anomaly, corrupting one glass panel and keeping the other clean. The glasses are mainly steady, with small variations in the angles with respect to the camera axes, which enrich the variability of the dataset. Moreover, the filters can be easily replaced when too dirty, making the whole data-collection process very fast. In contrast to Porav \textit{et al.}~\cite{porav2019can}, our device can be hand-held, which adds flexibility and allows us to collect data in any area of interest.
\begin{figure*}[t]
\begin{center}
    \includegraphics[scale=0.532]{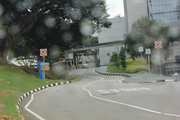} % first figure itself
    \includegraphics[scale=0.532]{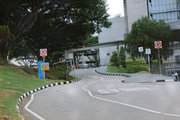} % first figure itself
    \includegraphics[scale=0.532]{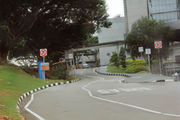} % first figure itself
    \includegraphics[scale=0.532]{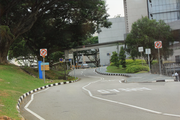} % first figure itself
    \includegraphics[scale=0.532]{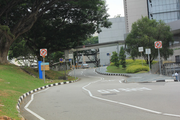} % first figure itself
    %\vspace{0.1cm}
    %\subcaptionbox{Medium task\label{fig:image1}}
    %\includegraphics[scale=0.13]{img4/p_10.jpeg} % first figure itself
    \includegraphics[scale=0.532]{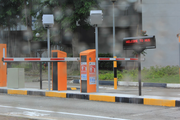} % first figure itself
    \includegraphics[scale=0.532]{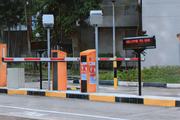} % first figure itself
    \includegraphics[scale=0.532]{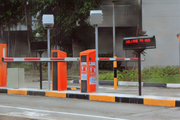} % first figure itself
    \includegraphics[scale=0.532]{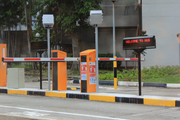} % first figure itself
    \includegraphics[scale=0.532]{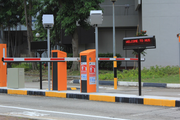} % first figure itself
    %\vspace{0.1cm}
    %\subcaptionbox{Difficult task\label{fig:image1}}
    %\includegraphics[scale=0.13]{img4/joint13_10.jpeg} % first figure itself
    \subcaptionbox{Input\label{fig:input}}
    {\includegraphics[scale=0.532]{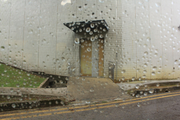}} % first figure itself
    \subcaptionbox{G\label{fig:G}}
    {\includegraphics[scale=0.532]{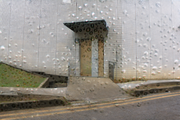}} % first figure itself
    \subcaptionbox{G+E\label{fig:GE}}
    {\includegraphics[scale=0.532]{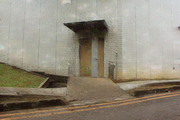}} % first figure itself
    \subcaptionbox{G+E+A\label{fig:GEA}}
    {\includegraphics[scale=0.532]{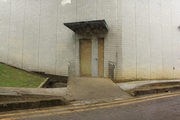}} % first figure itself
    \subcaptionbox{Ground Truth\label{fig:GT}}
    {\includegraphics[scale=0.532]{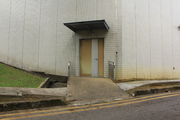}} % first figure itself
\end{center}
    \vspace{-0.3cm}
    \caption{Outputs of the different stages of our architecture. Rows correspond to easy, medium or difficult task, depending on how raindrops affect the images.}
    \label{img:diff_stages}
\end{figure*}
\begin{figure*}[t]
\begin{center}
    \includegraphics[scale=0.266]{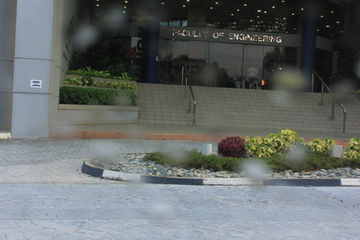} % first figure itself
    \includegraphics[scale=0.266]{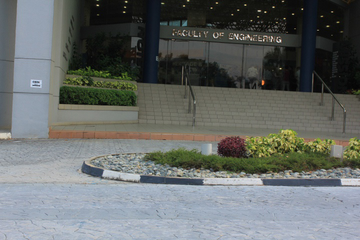} % first figure itself
    \includegraphics[scale=0.266]{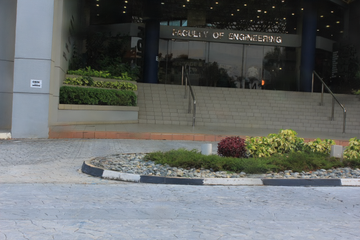} % first figure itself
    \includegraphics[scale=0.266]{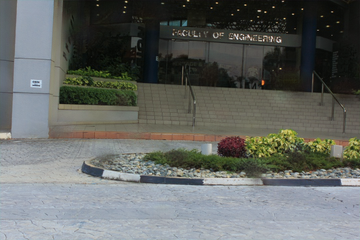} % first figure itself
    \includegraphics[scale=0.266]{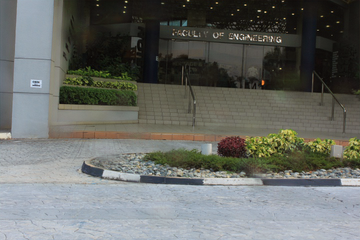} % first figure itself
    %\vspace{0.1cm}
    %\subcaptionbox{Medium task\label{fig:image1}}
    \includegraphics[scale=0.266]{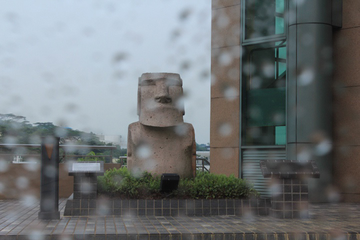} % first figure itself
    \includegraphics[scale=0.266]{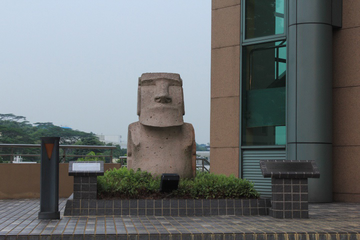} % first figure itself
    \includegraphics[scale=0.266]{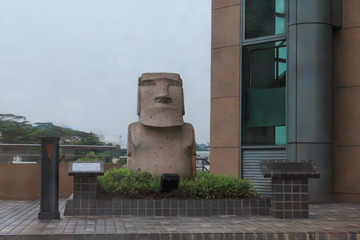} % first figure itself
    \includegraphics[scale=0.266]{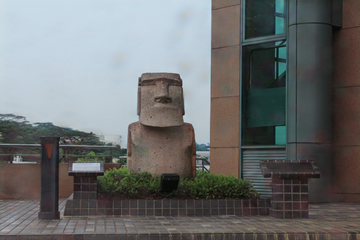} % first figure itself
    \includegraphics[scale=0.266]{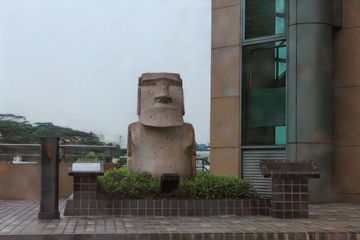} % first figure itself
    %\vspace{0.1cm}
    %\subcaptionbox{Difficult task\label{fig:image1}}
    \subcaptionbox{Input\label{fig:image1}}
    {\includegraphics[scale=0.266]{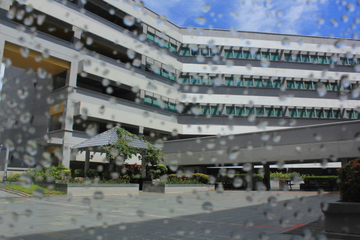}} % first figure itself
    \subcaptionbox{Ground Truth\label{fig:image1}}
    {\includegraphics[scale=0.266]{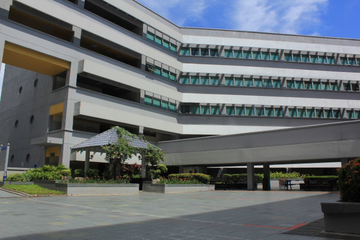}} % first figure itself
    \subcaptionbox{DeRaindrop~\cite{qian2018attentive}\label{fig:image1}}
    {\includegraphics[scale=0.266]{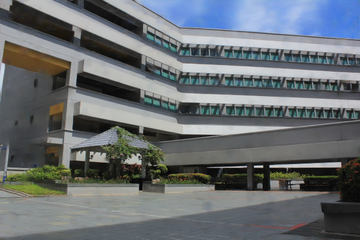}} % first figure itself
    \subcaptionbox{Porav \textit{et al.}~\cite{porav2019can}\label{fig:image1}}
    {\includegraphics[scale=0.266]{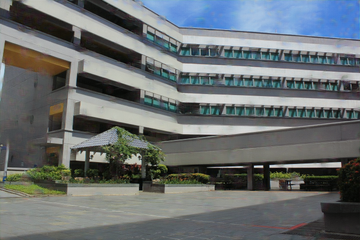}} % first figure itself
    \subcaptionbox{Ours\label{fig:image1}}
    {\includegraphics[scale=0.266]{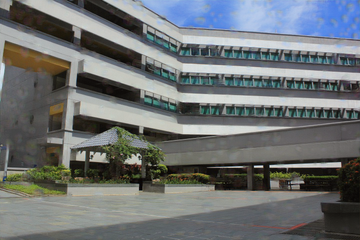}} % first figure itself
\end{center}
    \vspace{-0.2cm}
    \caption{Visual comparison between our results and the state of the art. Samples from the dataset by Qian \textit{et al.}~\cite{qian2018attentive}}
    \vspace{-0.6cm}
    \label{img:sota}
\end{figure*}
With this sensor, we collected 5613 image pairs of size 540x360 (after cropping) walking around the cities of Zurich and Milan, at different times of the day and with several weather conditions, with the aim to increase the variation of the scene. The majority of the samples are spoiled by water sprayed directly on the glass, mimicking raindrops with diverse shapes and sizes. Furthermore, we interchanged between left and right camera for the ground truth image due to a slight difference in illumination levels between the two, which automatically makes the training on this dataset more robust to this intrinsic noise. The images are rectified to reproduce the same scene, except for the occluded regions. In Figure~\ref{img:dataset} some samples from our dataset are displayed. Our device records 1-channel images, which can be used for many robotics tasks, \textit{e.g} SLAM and obstacle detection.

\subsection{Training Details}
We use the standard training scheme for image-to-image translation problems~\cite{isola2017image}. The discriminator is trained on a previously de-rained image and then the generator and the enhancer are updated in one step, minimizing our objective function, in contrast with Qu \textit{et al.}~\cite{qu2019enhanced}, where a dedicated training scheme is designed to incorporate the enhancer. We adopt the Adam~\cite{kingma2014adam} optimizer with learning rate $\lambda = 0.0002$, batch size of 8 and train for 200 epochs, using early stopping on validation set with patience 10. Experiments have been performed on an Nvidia RTX 2080 GPU and an Intel Core i7 10th Gen CPU.

\subsection{Ablation Study}
In this section, we state dissimilarities and improvements across incremental configurations of our architecture. We will consider three variants, tested on the dataset provided by Qian \textit{et al.}: 
\begin{itemize}
    \item G is the output of the generator, without enhancement and aggregation scheme;
    \item G + E is the image produced by the combination of enhancer module and generator;
    \item G + E + A incorporates the aggregation scheme.
\end{itemize}
\vspace{0.05cm}
Visual results can be found in Figure~\ref{img:diff_stages}. For the easy task -- top row -- where the input image is faintly affected by raindrops, G is sufficient to perform the anomaly mitigation. On the medium task -- central row -- instead, brighter spots can be clearly distinguished on the right wall of G, as clear signs of attempts to remove raindrops. On the other hand, G+E, although slightly over-colored, is able to produce adequate results, because of the input information directly injected into the enhancer. Nevertheless, it fails on the hard task -- bottom row -- where the raindrops introduce too much noise into the input image and, consequently, into the enhancer. In this case, the counter-effect produced by the generator needs to be stronger. With the introduction of the aggregation scheme (G+E+A), the network leverages feature maps learned at previous stages to retain relevant information and output successfully de-rained images.

From a quantitative point of view, as reported in Table~\ref{Tab:diff_stages}, the visual results are confirmed by the changes in Structural SIMilarity (SSIM)~\cite{wang2004image} and Peak Signal-To-Noise Ratio (PSNR)~\cite{fardo2016formal}, two widely-accepted metrics for this task. The enhancer module leads to the highest delta in both metrics, stating the importance of input information to output realistic images. The aggregation scheme further improves the results by providing the generator with more power to balance input noise introduced into the enhancer by heavily corrupted samples.
\begin{table}[t]
    \centering
    \begin{tabular}[b]{ l|c|c}
        \hline
        \noalign{\smallskip}
        Method & SSIM & PSNR\\
        \noalign{\smallskip}
        \hline
        \noalign{\smallskip}
        Input & 0.851 & 24.09\\ 
        G & 0.869 & 27.12\\
        G+E & 0.898 & 29.15\\
        G+E+A &  \textbf{0.909} &  \textbf{29.84}\\
        \noalign{\smallskip}
        \hline
    \end{tabular}
    \vspace{0.05cm}
    \caption{Results on the dataset by Qian \textit{et al.}~\cite{qian2018attentive} produced by different stages of our architecture. The experiments are run with synthetic initialization.}
    \vspace{-0.1cm}
\label{Tab:diff_stages}
\end{table}
\begin{table}[t]
    \centering
    \begin{tabular}[b]{ l|c|c}
        \hline
        \noalign{\smallskip}
        Initialization & SSIM & PSNR\\
        \noalign{\smallskip}
        \hline
        \noalign{\smallskip}
        Raw & 0.760 & 19.34\\
        \hline
        \noalign{\smallskip}
        Random & 0.787 & 22.36\\ 
        Synthetic~\cite{porav2019can} & 0.791 & 22.15\\
        Our synthetic & \textbf{0.813} & \textbf{24.77}\\
        \noalign{\smallskip}
        \hline
    \end{tabular}
    \vspace{0.05cm}
    \caption{Effect of different initialization strategies on our full architecture.}
    \vspace{-0.4cm}
\label{Tab:syn}
\end{table}
\subsection{Impact of synthetic data}\label{exp_syn}
To investigate how realistic and effective our synthetic raindrops are, we corrupt images from existing datasets (\cite{qian2018attentive, porav2019can}) and use them to pre-train our architecture. Then, we initialize the training on real data with the weights learned during the pre-training phase. We compared this setup with the same procedure applied to the synthetic dataset provided by Porav \textit{et al.}~\cite{porav2019can} and with a random initialization drawn from a Gaussian distribution with $\mu = 0$ and $\sigma = 0.02$. Results are shown in Table~\ref{Tab:syn}. Our synthetic initialization outperformed both the others, proving on one hand that artificial anomalies can be useful for this task by improving the random initialization; on the other hand that ours are more realistic with respect to our reference model.
\subsection{Comparison with state of the art}
Figure~\ref{img:sota} shows visual outputs from our architecture compared to the state of the art. Again, we distinguish between easy, medium and hard task. In the first two rows, all reconstructed images present little or no difference between each other and the ground truth. Only in the hard task - bottom row - looking carefully, the output of DeRaindrop looks cleaner and with less artifacts. This slight difference is reflected in the metrics, as depicted in Table~\ref{Tab:sota}. Our model ranks second in SSIM and third in PSNR. However, our architecture outperforms the others by a great margin in terms of average time required to de-noise the images. With just \emph{6 milliseconds} needed to run on the GPU, we are 15 to 40 times faster then the other works. And 5 to 8 times faster on the CPU, where it takes around half a second to run. This means that our network is the most suitable for a practical application on real robots, as it represents a good trade-off between speed and effectiveness. 
%\begin{itemize}
%    \item Random, i.e. drawn from a Gaussian distribution with mean 0 and variance 0.02;
%    \item From weights learned on the synthetic dataset provided by Porav \textit{et al.}~\cite{porav2019can};
%    \item From weights learned on 
%\end{itemize} 

%\subsection{Quantitative Results}
%\begin{itemize}
%    \item Reconstruction on my dataset for different modules (with/without enhancer, with/without skip %connections, with/without dilation etc.) explaining why we obtain those results;
%    \item Impact of synthetic data and explain why they work/don't work;
%    \item State-of the art comparison with metrics SSIM, PSNR and inference time;
%    \item If we have time we can add the impact of reconstruction on another task with synthetic raindrops %applied.
%\end{itemize}
%\subsection{Qualitative Results}
%\begin{itemize}
%    \item Images of results of our model with different configurations as in first point of quantitative %evaluation;
%    \item Comparison of images resulting from our model against other approaches (optional);
%    \item Visual results from a different task with applied synthetic raindrops with and without reconstruction %(can be very catchy).
%\end{itemize}
%\clearpage
%\ImageBox{7cm}{17cm}{Mine vs their synthetic drops}\\\\
%\\\\
%\ImageBox{7cm}{17cm}{Reconstructions on my dataset}\\\\
%\\\\
 
%\clearpage
%\ImageBox{10cm}{17cm}{Tables}\\\\
%\\\\
\begin{table}[t]
    \centering
    \begin{tabular}[b]{ l|c|c|c|c}
        \hline
        \noalign{\smallskip}
        Method & SSIM & PSNR & TIME GPU(s) & TIME CPU(s)\\
        \noalign{\smallskip}
        \hline
        \noalign{\smallskip}
        DeRaindrop~\cite{qian2018attentive} & \textbf{0.921} & 31.50 & 0.262 & 4.317\\ 
        Porav \textit{et al.}~\cite{porav2019can} & 0.902 & \textbf{31.55} & 0.091 & 2.780\\
        Ours & 0.909 & 29.84 & \textbf{0.006} & \textbf{0.542}\\
        \noalign{\smallskip}
        \hline
    \end{tabular}
    \vspace{0.05cm}
    \caption{Comparison with the state of the art. Our architecture largely outperforms the others in terms of time taken for a forward pass, while still producing comparable results.}
    \vspace{-0.5cm}
\label{Tab:sota}
\end{table}
\vspace{0.1cm}
\section{CONCLUSION}
%\begin{itemize}
%    \item Usually quite short in ICRA;
%    \item Summarize contributions;
%    \item Talk briefly about future works;
%    \item Add link to video/dataset.
%\end{itemize}
In this work we have presented our framework for fast camera-anomaly deletion and image restoration. We added aggregation modules and an enhancing block to a shallow GAN architecture to preserve state-of-the-art performance, while achieving real-time capabilities, which is our main focus. The proposed method can be run as a pre-processing step in every pipeline that involves classification, segmentation or localization tasks for autonomous robots in challenging conditions, where rain or dirt could spoil the camera lens and degrade performance. Our pipeline is suitable to be combined with other architectures in a real-time fashion, also considering the limited computing power of autonomous robots. 
In the future, we will inspect other anomalies, e.g. scratches, dust and fingerprints and will combine our architecture with a classification pipeline. This way, the robot could distinguish whether the camera is clean or affected by anomalies and, if that is the case, run it through the mitigation pipeline. Another classification stage would decide which anomaly is corrupting the images and leverage the corresponding pre-trained model.
\clearpage
{
\bibliographystyle{ieee_fullname}
\bibliography{references}
}

\end{document}